\documentclass[10pt,twocolumn,letterpaper]{article}

\usepackage{cvpr}              
\definecolor{cvprblue}{rgb}{0.21,0.49,0.74}
\usepackage[pagebackref,breaklinks,colorlinks,allcolors=cvprblue]{hyperref}
\usepackage{url}            
\usepackage{booktabs}       
\usepackage{amsfonts}       
\usepackage{nicefrac}       
\usepackage{microtype}      
\usepackage{xcolor}         
\usepackage{colortbl}
\usepackage{algpseudocode}
\usepackage{amsmath}
\usepackage{amssymb}
\usepackage{algorithm}
\usepackage{graphicx}
\usepackage{multirow}
\usepackage{siunitx}
\usepackage{colortbl}
\usepackage{subcaption}
\usepackage{siunitx}
\usepackage{cuted}
\usepackage{subcaption}
\usepackage[table]{xcolor}

\def\confName{CVPR}
\def\confYear{2026}

\title{CRAFT-LoRA: Content-Style Personalization via \\ Rank-Constrained Adaptation and Training-Free Fusion}

\author{Yu Li\textsuperscript{1,2}
\quad
Yujun Cai\textsuperscript{3} \quad
Chi Zhang\textsuperscript{1}\thanks{Corresponding author.}\\
\textsuperscript{1}AGI Lab, Westlake University \quad
\textsuperscript{2}George Washington University  \quad
\textsuperscript{3}The University of Queensland\\
{\tt\small yul@gwu.edu, chizhang@westlake.edu.cn}
}

\newcommand{\n}{\textbf{CRAFT-LoRA}}

\makeatletter
\def\abstract{%
   \iftoggle{cvprpagenumbers}{}{%
     \thispagestyle{empty}
   }
   \centerline{\large\bf Abstract}%
   \vspace*{2pt}\noindent%
   \it\ignorespaces%
}
\makeatother

\begin{document}
\twocolumn[{%
\maketitle
\vspace*{-5mm}
\centering
\includegraphics[width=\textwidth]{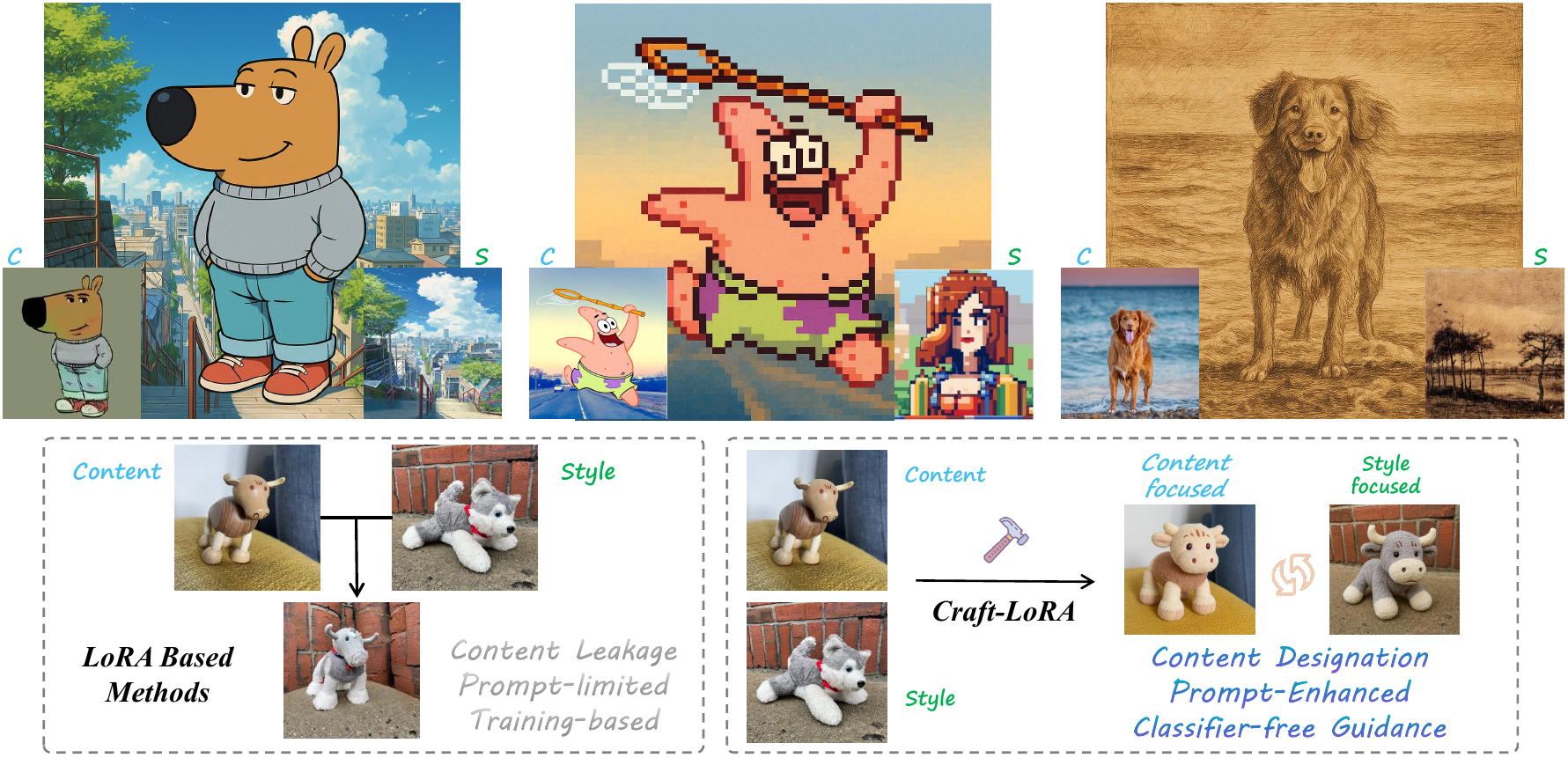}
\captionof{figure}{Samples generated by our proposed framework. Our method achieves more effective decoupling and fusion of content and style, enabling finer control over both aspects during generation.}
\label{fig:teaser}
\vspace*{3mm}
}]

\renewcommand{\thefootnote}{\fnsymbol{footnote}}
\footnotetext[1]{Corresponding author.}

\begin{abstract}
Personalized image generation requires effectively balancing content fidelity with stylistic consistency when synthesizing images based on text and reference examples. Low-Rank Adaptation (LoRA) offers an efficient personalization approach, with potential for precise control through combining LoRA weights on different concepts. However, existing combination techniques face persistent challenges: entanglement between content and style representations, insufficient guidance for controlling elements' influence, and unstable weight fusion that often require additional training. We address these limitations through \n, with complementary components: (1) rank-constrained backbone fine-tuning that injects low-rank projection residuals to encourage learning decoupled content and style subspaces; (2) a prompt-guided approach featuring an expert encoder with specialized branches that enables semantic extension and precise control through selective adapter aggregation; and (3) a training-free, timestep-dependent classifier-free guidance scheme that enhances generation stability by strategically adjusting noise predictions across diffusion steps. Our method significantly improves content-style disentanglement, enables flexible semantic control over LoRA module combinations, and achieves high-fidelity generation without additional retraining overhead. Code is avalible at \href{https://github.com/Skylanding/CraftLoRA}{https://github.com/Skylanding/CraftLoRA}.
\end{abstract}

\section{Introduction}
\label{sec:intro}

Recent advancements in deep learning have fueled remarkable progress in generative models, particularly within personalized image generation~\cite{ghebrehiwet2024revolutionizing,bengesi2024advancements,sengar2024generative}.
Unlike generic image synthesis, personalized generation focuses on customizing outputs based on user-provided references, such as a specific subject's appearance or stylistic attributes, enabling applications in creative design, entertainment, digital avatars, and personalized marketing.

Among the various techniques developed for personalized image generation, Low-Rank Adaptation (LoRA)~\cite{hu2022lora} has emerged as a particularly efficient approach by introducing small, trainable low-rank matrices into pre-trained text-to-image diffusion models, allowing fine-tuning on new concepts using only a small number of references. 
When integrated with frameworks like DreamBooth, LoRA-based methods enable customization to specific content or style from minimal examples~\cite{ryu2023low}. Moreover, multiple independently trained LoRA modules can be simultaneously applied during inference, enabling images that blend diverse attributes such as combining a specific subject with a distinct artistic style. 
However, naively merging multiple LoRA modules often leads to degraded quality or semantic inconsistency due to entangled representations, motivating recent work on more effective fusion strategies~\cite{shah2024ziplora,frenkel2024implicit,ouyang2025k}.

Despite recent progress, existing LoRA combination strategies still face several fundamental challenges. First, although pre-trained diffusion models empirically exhibit some degree of composability with respect to LoRA modules, these models are not explicitly trained to support such combinations. 
As a result, directly merging LoRA weights often fails to achieve clean disentanglement between content and style factors~\cite{liu2022learning,liu2024unziplora,avrahami2023break}. 
Inspired by the meta-learning paradigm, particularly MAML~\cite{finn2017model,chua2021fine}, which seeks optimal initializations for fast and generalizable task adaptation, we introduce a low-rank constrained adaptation phase inspired by PaRa~\cite{chen2024personalizing} that narrows the model's generative space to a more compact and disentangled representation that better supports multi-LoRA fusion.

Second, the notion of a subject in personalized image generation often encompasses a hierarchy of visual attributes, \eg facial identity, hairstyle, clothing style, and accessory details. However, current methods typically collapse this rich semantic structure into a single coarse-grained token representation~\cite{pavllo2020controlling,liu2023more}, neglecting the internal structure of subject attributes and lacking mechanisms to control the preservation of fine-grained elements. 
To address this, we propose an expert encoder system with specially designed content and style adapters operating in disjoint network layers. The expert encoder processes identity labels, content descriptions, and style specifications through separate branches, facilitating selective aggregation based on semantic relevance and enabling fine-grained control over feature preservation at different levels.

Third, existing LoRA fusion strategies such as ZipLoRA~\cite{shah2024ziplora} typically rely on additional optimization procedures to reconcile differences between LoRA modules~\cite{wang2024parameter,han2024parameter}. Directly modifying LoRA parameters during fusion can inadvertently alter or suppress critical elements encoded in the original weights, leading to loss of identity or stylistic fidelity, while also incurring non-trivial computational overhead. To overcome these limitations, we propose a tuning-free asymmetric Classifier-Free Guidance (CFG) mechanism. Rather than adjusting LoRA weights directly, we selectively apply time-dependent modifications only to the conditional path while keeping the unconditional path anchored to base model weights, yielding a dynamic guidance signal that steers generation toward a desirable content-style balance without additional training cost.

\begin{figure*}[!t]
        \centering  \includegraphics[width=1.0\textwidth]{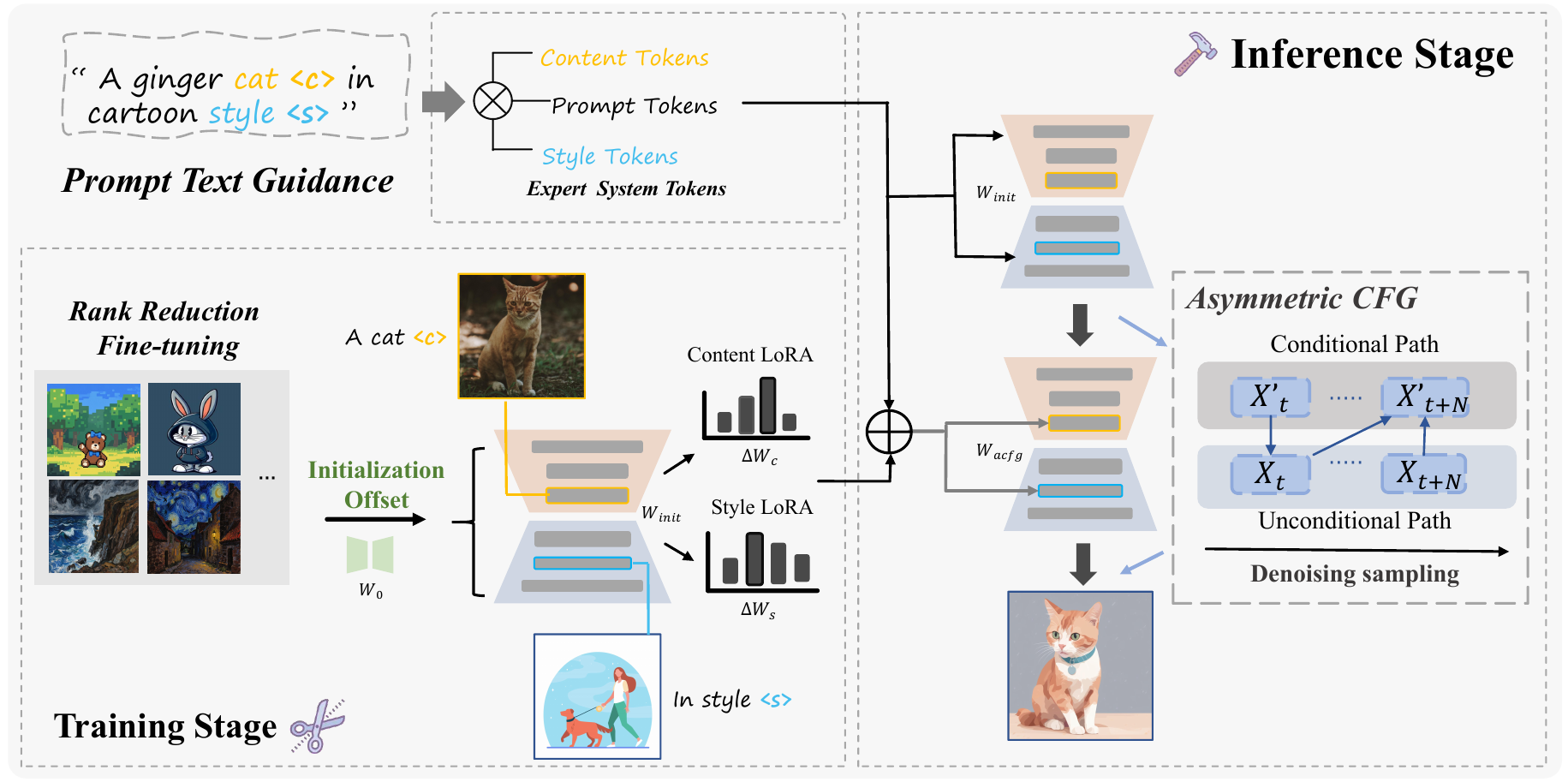}
    \caption{Overview of \textbf{\n}, a unified pipeline for personalized image synthesis. In the \emph{Training Stage}, content ($\Delta W_c$) and style ($\Delta W_s$) LoRA adapters are decoupled from reference images using a rank-restricted initialization offset. The \emph{Prompt Text Guidance} module employs an expert system with specialized branches to produce distinct content and style tokens for semantic control. In the \emph{Inference Stage}, a timestep-aware asymmetric CFG scheme selectively integrates LoRA updates, ensuring stable and high-fidelity image generation.
}
    \label{fig:model}
\end{figure*}

As illustrated in Figure~\ref{fig:model}, our proposed unified framework integrates rank-constrained fine-tuning to learn decoupled content and style subspaces, leverages prompt-guided mechanisms for semantic control via selective adapter aggregation guided by an expert encoder, and employs a timestep-aware guidance correction scheme for stable, high-fidelity generation without extra training.

To summarize, we make the following key contributions:
\begin{itemize}
    \item We propose a novel framework that enhances content-style disentanglement during LoRA training through rank-constrained fine-tuning and low-rank projection residuals, encouraging the learning of decoupled subspaces.
    \item We introduce a prompt-guided approach with an expert encoder and selective adapter aggregation, enabling precise semantic control over content and style influences and extending the applicability of LoRA modules.
    \item We develop a training-free time-step-aware and classifier-free guidance correction scheme that improves the stability and fidelity of diffusion-based generation by strategically adjusting noise predictions during critical steps.
\end{itemize}

\section{Related Work}
\label{sec:rw}

\subsection{Personalized image generation}
The research trajectory of image stylization has evolved from early Generative Adversarial Network-based domain translation methods—often relying on paired or unpaired datasets and extensive, domain-specific training—to diffusion-based frameworks~\cite{cai2021generative,rombach2022high,blattmann2023align}. More recently, the application of large pre-trained vision–language models has further advanced zero-shot editing and stylization. For instance, Prompt-to-Prompt~\cite{hertz2022prompt} guides the generation process by selectively modulating cross-attention maps, while plug-and-play methods introduce lightweight adapters into self-attention layers to adjust spatial features dynamically~\cite{ma2025scflow,hao2023vico,zhang2024pia}.
Despite significant progress, reconciling high content fidelity with flexible style injection remains an open challenge~\cite{chung2024style,li2024beyond}. Practice shows that naive parameter fusion strategies, such as weighted averaging, often lead to unintended mixing of content and style features~\cite{shuai2024survey}. Consequently, current stylization pipelines often struggle to integrate diverse visual concepts without compromising either semantic coherence or stylistic integrity~~\cite{tanveer2023ds,park2023visual}.

\subsection{Fine-Tuning and LoRA Composition for Stylization}
The rapid advancement of text-to-image diffusion models has spurred diverse personalization strategies. Full-model fine-tuning approaches such as DreamBooth~\cite{ruiz2023dreambooth} achieve high-fidelity transfer but incur substantial computational costs and require lengthy retraining on domain-specific examples. To mitigate these demands, Parameter-Efficient Fine-Tuning (PEFT) techniques have emerged~\cite{ruiz2024hyperdreambooth,han2024parameter}. Among these, LoRA~\cite{hu2022lora} introduces small, trainable low-rank decomposition matrices, injecting new stylistic or semantic concepts while keeping the majority of pre-trained weights frozen, thus striking a balance between adaptability and efficiency. Complementary approaches include token-based optimization techniques such as Textual Inversion~\cite{gal2022image}, which learns new embeddings to represent novel concepts within the model's token vocabulary, and adapter-based schemes such as StyleDrop~\cite{sohn2023styledrop}, which appends lightweight layers to attention blocks to isolate style information in dedicated modules. 
Meta-learning, as a cross-task method for domain transfer, shares similar principles with style transfer in personalized image generation, providing new inspirations for fine-tuning methods~\cite{zhao2022effectiveness}. 

A key challenge arises when combining multiple LoRA modules for simultaneous content and style adaptation. Parameter interpolation~\cite{tang2023parameter,wang2024lora} provides a basic approach but is prone to suboptimal outcomes, especially when parameters conflict. To address this, several studies refine the composition mechanism: ZipLoRA~\cite{shah2024ziplora} learns column-wise mixing coefficients for each content-style pairing, B-LoRA~\cite{frenkel2024implicit} differentiates functional roles within attention modules for content-style separation, and LoRA Composition~\cite{zhong2024multi} employs a cyclic update mechanism to iteratively guide generation without explicit parameter merging. Training-free approaches have also been explored, with K-LoRA~\cite{ouyang2025k} selectively applying LoRAs to different attention layers and CLoRA~\cite{meral2025contrastive} proposing contrastive test-time composition. Other directions include QR-LoRA~\cite{yang2025qr}, which applies QR decomposition for parameter-efficient training, and LoRA.rar~\cite{shenaj2025lora}, which learns merging via hypernetworks. While these methods improve the merging or composition process, the upstream entanglement inherent in standard backbones remains underexplored.

\section{Methodology}
\label{sec:method}
\subsection{Preliminaries}
LoRA is an efficient fine-tuning method that adapts pre-trained diffusion models by adding trainable low-rank matrices $A \in \mathbb{R}^{r \times n}$ and $B \in \mathbb{R}^{m \times r}$ to frozen weights $W_0 \in \mathbb{R}^{m \times n}$, yielding $W = W_0 + BA$ with $r \ll \min(m,n)$.
In personalized image generation, content LoRAs $\Delta W_c$ represent subject-specific features such as identity and structure, while style LoRAs $\Delta W_s$ capture artistic aspects such as texture and color.
Both can be trained efficiently from a single reference image, and during inference they can be composed into the backbone to produce personalized outputs.

\subsection{Framework Overview}
We propose \n, a unified framework for personalized image synthesis that addresses content-style disentanglement and fusion in LoRA-based methods. As shown in Fig.~\ref{fig:model}, the framework consists of three main stages:
(1) \textit{Training Stage}, where content and style LoRA weights are learned separately based on a rank-constrained fine-tuned backbone;  
(2) \textit{Prompt Text Guidance}, which employs an expert encoder system with specialized branches to provide semantic control over content and style during training and inference; 
(3) \textit{Inference Stage}, utilizing a timestep-aware asymmetric classifier-free guidance scheme for stable and flexible fusion of LoRA weights without extra training.

\subsection{Rank-Limited Fine-Tuning of Backbone}
As discussed in the introduction, pre-trained diffusion models do not explicitly support clear disentanglement of content and style factors when adapting weights from a reference image. Motivated by the adaptability of MAML~\cite{finn2017model,li2025dual} under few-shot constraints and leveraging the rich hierarchical features of SDXL~\cite{podell2023sdxl}, we propose to fine-tune the U-Net backbone under a rank-constrained mechanism that provides optimized initialization and structurally improves content-style disentanglement for subsequent LoRA training.

\textbf{Rank Reduction Fine-Tuning.}
Inspired by PaRa~\cite{chen2024personalizing}, we denote the frozen backbone weight matrix of layer $l$ as $W_l^{(0)} \in \mathbb{R}^{d_{\mathrm{out}}^l \times d_{\mathrm{in}}^l}$, with $L$ layers in total. We introduce a learnable basis matrix $B_l \in \mathbb{R}^{d_{\mathrm{in}}^l \times r_l}$ per layer, where $r_l$ is the rank constraint satisfying $r_l \ll \min(d_{\mathrm{out}}^l, d_{\mathrm{in}}^l)$. Computing the QR decomposition $B_l = Q_l R_l$ yields an orthonormal basis $Q_l$ with $Q_l^\top Q_l = I$.
The rank-limited fine-tuning projects out components of \( W_l^{(0)} \) along \( Q_l \), yielding updated weights:
\begin{equation}
W_l = W_l^{(0)} - Q_l Q_l^\top W_l^{(0)}.
\end{equation}
This operation constrains updates to directions orthogonal to the learned low-rank subspace, introducing an architectural bias that encourages reduced overlap between content and style representations. We show empirically in Appendix~\ref{sec:appendix} that it reduces cross-influence.

To further allocate adaptation capacity hierarchically, we schedule:
\begin{equation}
r_l = r_{\max} - \frac{l-1}{L-1}(r_{\max} - r_{\min}), \quad l=1, \ldots, L,
\end{equation}
assigning higher ranks to earlier layers, which typically encode low-level structural and textural information where content and style factors are more intertwined, thus requiring more adaptation capacity. In implementation, we set $r_{\max}=128$ and $r_{\min}=4$. Compared to a uniform rank $r=64$ across all layers, this schedule improves Content Similarity by 4.8\%, Style Similarity by 5.2\%, and reduces cross-influence by 11.3\%, confirming the benefit of hierarchical rank allocation.

For content-style disentanglement, we train separate basis matrices \( B_l^{\mathrm{content}} \) and \( B_l^{\mathrm{style}} \), obtaining corresponding orthonormal subspaces \( Q_l^{\mathrm{content}} \) and \( Q_l^{\mathrm{style}} \). Merging these via concatenation and QR decomposition yields the combined weight update. Specifically, we first concatenate the basis matrices $[B_l^{\mathrm{content}} \mid B_l^{\mathrm{style}}]$ and apply QR decomposition to their corresponding orthonormal subspaces:
\begin{equation}
Q_l^{\mathrm{merged}} = [Q_l^{\mathrm{content}} \mid Q_l^{\mathrm{style}}] = Q_l^{\mathrm{combined}} R_l^{\mathrm{combined}}
\label{eq:rank1}
\end{equation}
\vspace{-\baselineskip}
\begin{equation}
W_l^{\mathrm{combined}} = W_l^{(0)} - Q_l^{\mathrm{combined}} (Q_l^{\mathrm{combined}})^\top W_l^{(0)}
\label{eq:rank2}
\end{equation}
The fine-tuned backbone weights are retained to provide a targeted initialization, thereby empirically reducing the cross-influence between content and style factors. After rank-limited fine-tuning, we denote the resulting backbone as 
\(W_{\text{init}} = \{W^{\text{init}}_i\}_{i=1}^L\), 
which serves as the host weights for all subsequent LoRA training and inference stages.

\begin{figure}[!t]
    \centering
    \includegraphics[width=0.9\columnwidth]{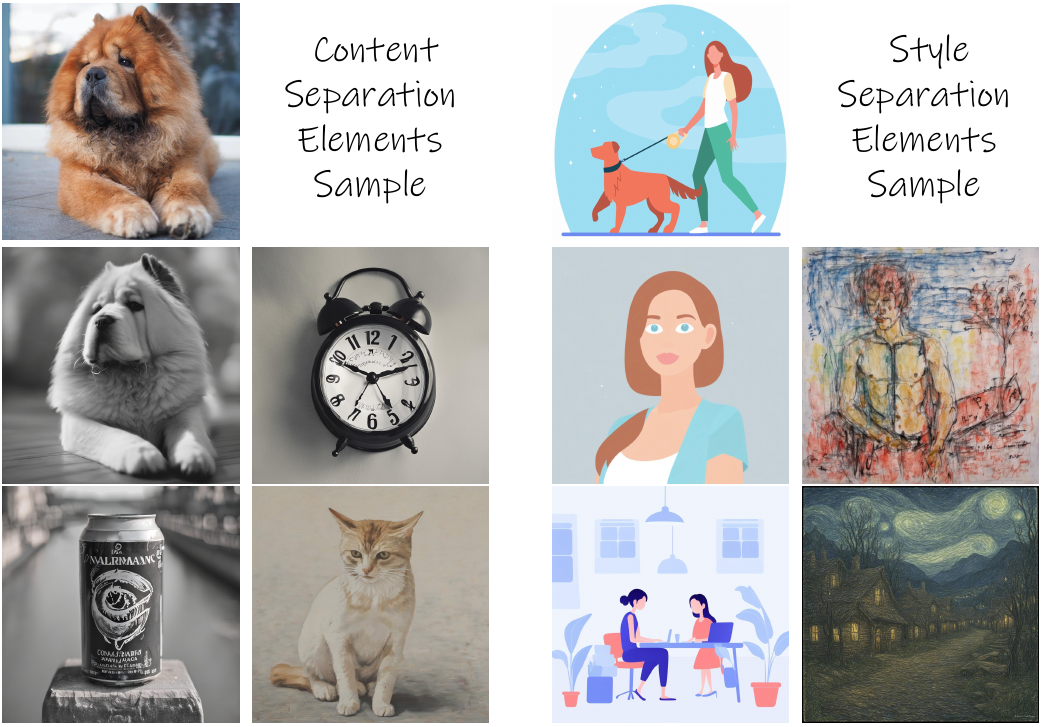}
    \caption{Content-style separation via frequency domain decomposition. Each group shows an original image with its frequency-separated elements and additional samples from the same group. Left: content group, where low-frequency components capture structural and semantic information. Right: style group, where high-frequency components capture textures and artistic rendering.}
    \label{fig:freq}
\end{figure}

\textbf{Contrastive Content-Style Pairs.}
To facilitate effective disentanglement during fine-tuning, we construct a curated dataset of $\mathbf{100}$ contrastive content-style image pairs. These pairs are designed so that either the content or the style remains constant while the other varies, guiding the model to separately capture content and style characteristics.

We operationalize the notion of content and style using frequency-domain decomposition: \textit{content} corresponds primarily to low-frequency components encoding structural and semantic information such as object layouts, while \textit{style} corresponds to high-frequency components capturing textures and color palettes.
Specifically, we apply Gaussian low-pass filtering with cutoff frequency $\sigma = 0.35$ (normalized) to extract content features, and compute the residual as style features. The cutoff was selected empirically by evaluating separation quality across $\sigma \in \{0.2, 0.25, 0.3, 0.35, 0.4, 0.45, 0.5\}$ via MLLM-based assessment, with $\sigma=0.35$ yielding the best content-style separation across diverse reference images.

The content and style samples under this frequency method are shown in Figure~\ref{fig:freq}.
Concretely, we select \(\mathbf{10}\) distinct content reference images \(\{C_i\}_{i=1}^{10}\) and \(\mathbf{10}\) distinct style reference images \(\{S_j\}_{j=1}^{10}\). Exhaustively pairing them yields:
\begin{equation}
\{ (C_i, S_j) \mid i=1,\dots,10;\; j=1,\dots,10 \}.
\end{equation}
Fixing a content image \(C_1\) and combining it with styles \(\{S_1,\dots,S_{10}\}\) produces 10 pairs that vary only in style, while fixing a style and varying content produces complementary pairs.

These contrastive pairs are directly used to drive the rank-limited backbone training described above. For each pair, the two images are fed as separate inputs: one provides supervision for learning the content-specific basis \(B_l^{\mathrm{content}}\), while the other supervises the style-specific basis \(B_l^{\mathrm{style}}\). The corresponding orthonormal subspaces \(Q_l^{\mathrm{content}}\) and \(Q_l^{\mathrm{style}}\) are then merged, and the backbone is updated according to Eqs.\ref{eq:rank1},\ref{eq:rank2}. 
In this way, the model allocates content and style variations to disjoint low-rank subspaces, ensuring that the backbone initialization $W_l^{\mathrm{combined}}$ encodes reduced interference between the two factors. 
We find that diversity matters more than quantity since 50 pairs already achieve 95\% of full performance, while increasing to 200 pairs yields less than 2\% additional improvement.

Empirically, combining rank-limited updates with contrastive pairs leads to a \textbf{12\%} reduction in average CLIP-based content–style similarity, indicating that the learned features exhibit reduced cross-influence. 
As further demonstrated in Table~\ref{fig:quantitative_data}(b), removing Rank-FT results in the largest performance drop among all components, quantitatively confirming its central role in disentanglement.
Details on dataset construction and training settings are provided in Appendix~\ref{sec:appendix}.

\begin{figure*}[htbp]
  \centering
  \begin{minipage}[t]{0.48\linewidth}
    \begin{algorithm}[H]
    \renewcommand{\algorithmicrequire}{\textbf{Input:}}
    \renewcommand{\algorithmicensure}{\textbf{Output:}}
      \caption{Prompt-Guided Decoupling \& Aggregation}
      \label{alg:prompted-lora}
      \begin{algorithmic}[1]
        \Require backbone \(W_{\text{init}}\); embeddings \(e_{\mathrm{sem}}, e_c, e_s\); content/style scalars \(\gamma_c,\gamma_s\)
        \Ensure aggregated backbone \(W^{\mathrm{agg}}\)
        \State Initialize updates \(\Delta W_i^{(c)} \gets 0, \Delta W_i^{(s)} \gets 0\) for all \(i\)
        \For{\(i = 1 \to L\)}
          \If{\(i \in I_c\)} update \(\Delta W_i^{(c)} \gets B_i A_i^{(c)}(e_{\mathrm{sem}})\)
          \ElsIf{\(i \in I_s\)} update \(\Delta W_i^{(s)} \gets B_i A_i^{(s)}(e_{\mathrm{sem}})\)
          \EndIf
        \EndFor
        \State Compute \(W^{\mathrm{agg}}\) as in Eq.~\eqref{eq:agg_injection}
        \State \Return \(W^{\mathrm{agg}}\)
      \end{algorithmic}
    \end{algorithm}
  \end{minipage}
  \hfill
  \begin{minipage}[t]{0.48\linewidth}
    \begin{algorithm}[H]
    \renewcommand{\algorithmicrequire}{\textbf{Input:}}
    \renewcommand{\algorithmicensure}{\textbf{Output:}}
      \caption{Asymmetric CFG (ACFG) Sampling}
      \label{alg:acfg}
      \begin{algorithmic}[1]
        \Require backbone \(W_{\text{init}}\); LoRA updates \(\{\Delta W_i^{(c)}\}, \{\Delta W_i^{(s)}\}\); schedules \(T_c,T_s\); guidance \(\omega\)
        \Ensure final sample \(\hat{x}_0\)
        \State Initialize \(x_T \sim \mathcal{N}(0,I)\); \textbf{for} \(t = T \to 1\) \textbf{do}
        \State \(\gamma_c(t) \gets \mathbb{1}_{t\in T_c},\ \gamma_s(t) \gets \mathbb{1}_{t\in T_s}\)
        \State Build \(W^{\mathrm{cond}}(t)\) as in Eq.~\eqref{eq:cond_weights}; set \(W^{\mathrm{uncond}} \gets W_{\text{init}}\)
        \State Compute \(\epsilon_{\mathrm{cond}}, \epsilon_{\mathrm{uncond}}\); form \(\epsilon_{\mathrm{acfg}}\) as in Eq.~\eqref{eq:acfg_prediction}
        \State \(x_{t-1} \gets \mathrm{Denoise}(x_t, \epsilon_{\mathrm{acfg}}, t)\)
        \State \textbf{end for}
        \State \Return \(\hat{x}_0\)
      \end{algorithmic}
    \end{algorithm}
  \end{minipage}
\end{figure*}

\subsection{Specialized Decoupling and Fusion}
Building on the rank-limited backbone initialization, we introduce a prompt-guided framework to further separate content and style at the adapter level. During training, content and style LoRA updates are learned on disjoint layer subsets with explicit prompt markers.
At inference, an expert encoder routes prompt information to selectively activate these adapters, enabling flexible fusion and fine-grained control.

\textbf{Prompt-Guided Decoupling and Encoding.}
Given a raw prompt $p$ containing explicit markers for content and style, \eg $p =$ ``A photo of a person \texttt{<c>} smiling in a watercolor style \texttt{<s>}'',
we treat these markers as routing signals to regulate LoRA module activation and to prevent leakage of content semantics into the style branch and vice versa. The prompt with markers removed is encoded by the SDXL text encoder to produce a unified semantic embedding:
\begin{equation}
e_{\mathrm{sem}} = \mathrm{SDXL\_Encoder}\bigl(p \setminus \{\texttt{<c>}, \texttt{</c>}, \texttt{<s>}, \texttt{</s>}\}\bigr),
\end{equation}
where \(e_{\mathrm{sem}} \in \mathbb{R}^d\) carries the overall semantics. Similarly, we extract content-specific embedding \(e_c\) and style-specific embedding \(e_s\) associated with \texttt{<c>} and \texttt{<s>} respectively, enabling isolated semantic guidance for each branch.

Following standard LoRA parametrization, each layer $i$ is updated by low-rank matrices. To bias the network towards disentanglement, we allocate disjoint index sets $I_c$ and $I_s$: lower and middle layers primarily capture structure and identity (content), while higher layers encode textures and rendering patterns (style).
Content-specific matrices \(A_i^{(c)}\) and style-specific matrices \(A_i^{(s)}\) are trained with the semantic embedding, and updates are restricted as
\begin{equation}
\Delta W_i = 
\begin{cases}
B_i\,A_i^{(c)}(e_{\mathrm{sem}}), & i \in I_c, \\
B_i\,A_i^{(s)}(e_{\mathrm{sem}}), & i \in I_s, \\
0, & \text{otherwise},
\end{cases}
\label{eq:decoupled_updates}
\end{equation}
so that gradient updates are confined to designated subsets guided by markers, which biases the learned representations towards more independent content and style subspaces. 

At inference, the Expert Encoder processes the prompt and produces control scalars \(\gamma_c,\gamma_s\) that regulate the strength of content and style adapters. By default, \(\gamma_c, \gamma_s \in \{0, 1\}\) are set based on marker presence, and users can also continuously adjust them within \([0, 1]\) for fine-grained control over content and style intensity. The aggregated backbone at inference is then computed as
\begin{equation}
W^{\mathrm{agg}}
= W_{\text{init}}
+ \sum_{i\in I_c}\mathsf{E}_i\!\left(\gamma_c\,\Delta W^{(c)}_i\right)
+ \sum_{i\in I_s}\mathsf{E}_i\!\left(\gamma_s\,\Delta W^{(s)}_i\right),
\label{eq:agg_injection}
\end{equation}
where \(W_{\text{init}}\) denotes the rank-limited backbone obtained in Eqs.\ref{eq:rank1},\ref{eq:rank2}. 
The operator $\mathsf{E}_i(\cdot)$ injects its input into the $i$-th layer while filling zeros elsewhere, so the summation operates as a layerwise direct sum.

If markers \texttt{<c>} or \texttt{<s>} are absent, the corresponding branch remains inactive. Users may also directly adjust $\gamma_c, \gamma_s$ to control the relative influence of content and style, enabling flexible compositions such as ``preserve content but replace style'' without retraining.
The full inference procedure is summarized in Algorithm~\ref{alg:prompted-lora}, showing how prompts route into disjoint LoRA branches and produce the aggregated backbone \(W_{\mathrm{agg}}\).

\textbf{Asymmetric Classifier-Free Guidance}
In standard classifier-free guidance (CFG), the conditional and unconditional paths share the same model weights. When LoRA updates are applied, this causes the unconditional path to be contaminated by content and style adapters, leading to instability in image generation. 
To address this issue, we propose Asymmetric Classifier-Free Guidance (ACFG), where the conditional path uses LoRA-adapted weights while the unconditional path remains anchored to the rank-limited backbone \(W_{\text{init}}\).

Let $i \in \{1, \ldots, L\}$ index model layers and $t \in \{1, \ldots, T\}$ diffusion timesteps. Instead of uniformly applying LoRA updates at all timesteps, we introduce time-dependent activation schedules $\gamma_c(t) = \mathbb{1}_{t \in T_c}$ and $\gamma_s(t) = \mathbb{1}_{t \in T_s}$, where $T_c$ and $T_s$ denote the active timestep ranges for content and style respectively. In practice, we activate content LoRA during early-to-mid timesteps to establish structural layout, and style LoRA during mid-to-late timesteps to refine textures and rendering, reflecting the coarse-to-fine nature of the diffusion process. The conditional weights at timestep $t$ are:
\begin{equation}
W^{\mathrm{cond}}(t) = W_{\text{init}}
+ \sum_{i\in I_c}\mathsf{E}_i\!\bigl(\gamma_c(t)\,\Delta W_i^{(c)}\bigr)
+ \sum_{i\in I_s}\mathsf{E}_i\!\bigl(\gamma_s(t)\,\Delta W_i^{(s)}\bigr),
\label{eq:cond_weights}
\end{equation}
while the unconditional path uses only the initialization backbone throughout:
\begin{equation}
W_i^{\mathrm{uncond}}(t) = W^{\text{init}}_i, \quad \forall\, t, i.
\label{eq:uncond_weights}
\end{equation}

The conditional and unconditional noise predictions are $\epsilon_{\mathrm{cond}} = \epsilon_{\theta}(x_t \mid W^{\mathrm{cond}}(t), \mathrm{cond})$ and $\epsilon_{\mathrm{uncond}} = \epsilon_{\theta}(x_t \mid W^{\mathrm{uncond}}, \varnothing)$ respectively. The final guided estimate is:
\begin{equation}
\epsilon_{\theta}^{\mathrm{acfg}} 
= (1+\omega)\,\epsilon_{\mathrm{cond}}
- \omega\,\epsilon_{\mathrm{uncond}},
\label{eq:acfg_prediction}
\end{equation}
where $\omega$ controls guidance strength. The key insight is that by anchoring the unconditional path to $W_{\text{init}}$, the guidance signal $\epsilon_{\mathrm{cond}} - \epsilon_{\mathrm{uncond}}$ isolates the effect of LoRA adapters at each timestep, rather than conflating adapter influence with the unconditional baseline.
Combined with the time-dependent schedules, this allows content structure to be established before style details are applied, reducing interference between the two factors. 
Crucially, ACFG requires no additional training or parameter optimization and maintains the same two-pass computation as standard CFG. Therefore, ACFG can be directly applied to standard SDXL LoRAs as a single module without the rank-limited backbone, providing improved fusion stability albeit with reduced disentanglement.

\begin{figure*}[!t]
    \centering
    \includegraphics[width=1.0\textwidth]{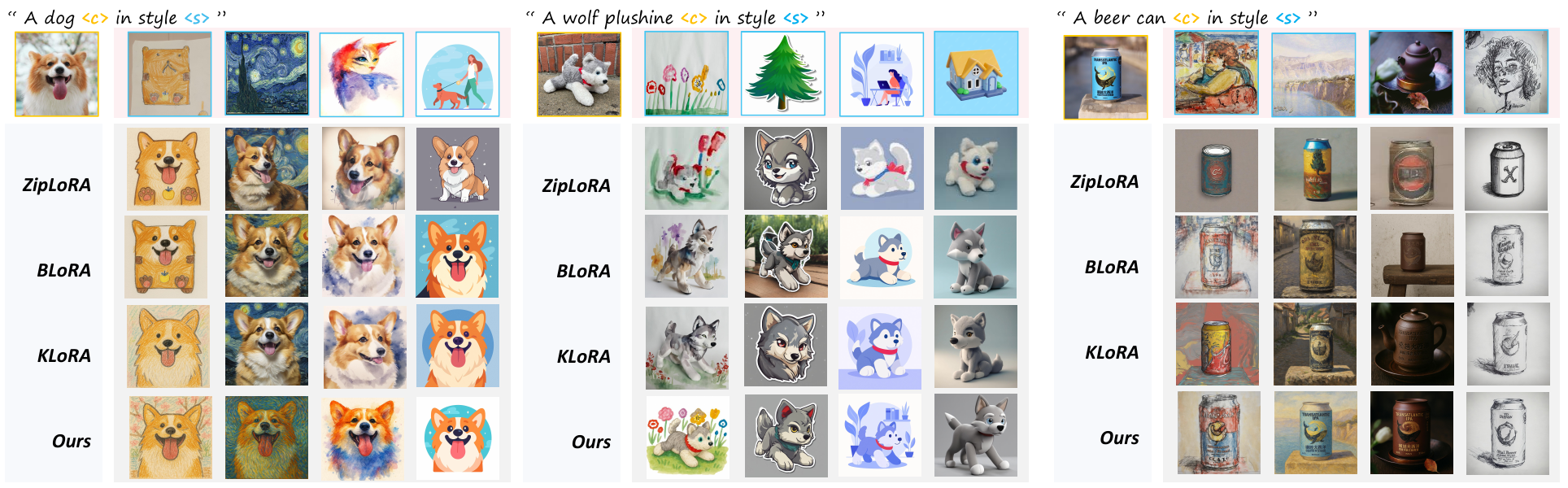}
    \caption{Visual comparison of content-style combinations. The figure presents a systematic evaluation of different methods in combining specific content elements with distinct artistic styles. Each column showcases the content and style reference followed by the outputs generated. Prompts follow the format ``A [content] \texttt{<c>} in [style] \texttt{<s>}''. Competing methods often fail to simultaneously preserve structure and render style, whereas our method produces consistent and coherent content–style compositions.}
    \label{fig:visual_results1}
\end{figure*}

\begin{figure}[!t]
    \centering
    \includegraphics[width=1.0\columnwidth]{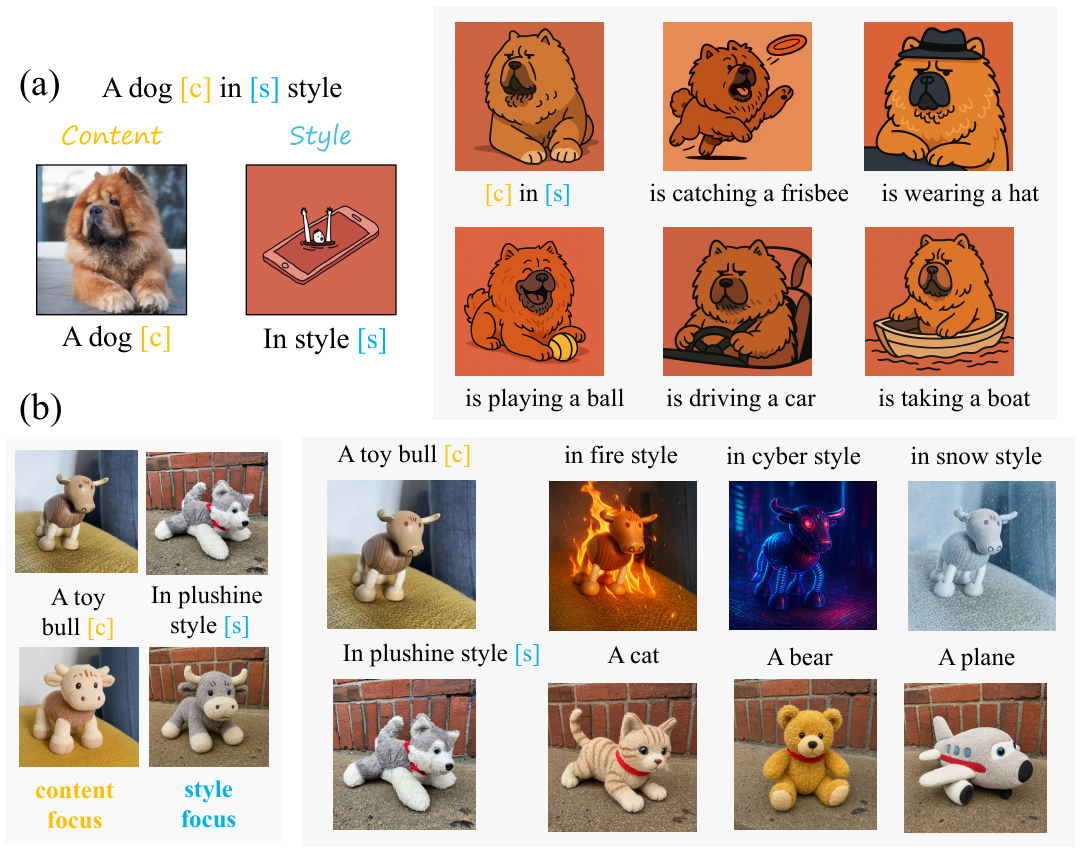}
    \caption{Extended visual results. 
    (a) Content–style generations augmented with additional prompt descriptions \eg ``catching a frisbee'', ``wearing a hat'', and ``driving a car''. Our method flexibly integrates dynamic semantics while preserving the specified content and style.
    (b) Single-branch generations where only the content LoRA or only the style LoRA is activated. Content-only generations preserve object identity across neutral renderings, while style-only generations transfer artistic features across different subjects.}
    \label{fig:visual_results2}
\end{figure}

\section{Experiments and Results}
\label{sec:result}
\subsection{Implementation Details}
For content concepts, we use the DreamBooth dataset~\cite{ruiz2023dreambooth} (30 themes, 4-5 images each). Style concepts come from StyleDrop~\cite{sohn2023styledrop}, supplemented with art style images. Each LoRA module is trained from a single reference image. We construct 100 contrastive pairs from these sources via frequency-domain decomposition to enhance content-style separation during backbone fine-tuning. 
For evaluation, we use strictly disjoint train/test splits. Each metric is averaged over 10 content-style pairs with 10 generated images per pair. Content and Style Similarity are computed using CLIP-I embeddings. The Combination Score follows MLLM-based binary judgment practices~\cite{lin2024evaluating}: GPT-4o is presented with the content reference, style reference, and output, and asked whether the output successfully integrates both content identity and artistic style. The final score is the average success rate over 50 evaluations per sample.

All experiments use Stable Diffusion XL~\cite{podell2023sdxl} on NVIDIA 4090 GPUs with DreamBooth training at rank $r=64$, Adam-8bit optimizer, 1000 steps, and learning rate \num{1e-5}. For rank-constrained backbone fine-tuning, we use $r_{\max}=128$, $r_{\min}=4$ with linear decay, assigning higher ranks to earlier layers where content and style are more entangled. Compared to uniform rank $r=64$, this schedule improves Content Similarity by 4.8\% and Style Similarity by 5.2\% while reducing cross-influence by 11.3\%. This fine-tuning is a one-time cost of 5000 steps on 2$\times$4090 GPUs, after which $W_{\text{init}}$ is reused for all subsequent LoRA training. During inference, ACFG maintains the same two-pass structure as standard CFG with less than 5\% additional overhead. For $T=50$ sampling steps, we set $T_c = [1, 35]$ and $T_s = [15, 50]$, with an overlapping window allowing content and style to jointly influence the mid-stage transition.

\begin{figure}[t]
    \captionsetup{type=table}
    \centering
    \begin{subfigure}[t]{1.0\linewidth}
        \centering
        \caption{Quantitative Comparison}
        \label{subfig:quant_comp_table}
        \resizebox{\linewidth}{!}{%
        \begin{tabular}{@{}lccc@{}}
            \toprule
            Method & \begin{tabular}[c]{@{}c@{}}Content Sim.\\{\small (CLIP-I)} $\uparrow$\end{tabular} & \begin{tabular}[c]{@{}c@{}}Style Sim.\\{\small (CLIP-I)} $\uparrow$\end{tabular} & \begin{tabular}[c]{@{}c@{}}Comb. Score\\{\small (GPT-4o)} $\uparrow$\end{tabular} \\
            \midrule
            Direct Merging        & 0.65 & 0.60 & 0.62 \\
            StyleDrop             & 0.68 & 0.76 & 0.70 \\ 
            ZipLoRA               & 0.70 & 0.69 & 0.73 \\
            KLoRA                 & 0.71 & 0.72 & 0.76 \\
            BLoRA                 & 0.74 & 0.70 & 0.77 \\
            LoRA.rar              & 0.73 & 0.71 & 0.76 \\
            \rowcolor{gray!15}
            \textbf{Ours}         & \textbf{0.79} & \textbf{0.80} & \textbf{0.83} \\
            \bottomrule
        \end{tabular}%
        }
    \end{subfigure}  
    \begin{subfigure}[t]{1.0\linewidth}
        \centering
        \caption{Component ablation study}
        \label{subfig:ablation_table}
        \resizebox{\linewidth}{!}{%
        \begin{tabular}{@{}ccc|ccc@{}}
            \toprule
            Rank-FT & Router & ACFG & Content Sim. $\uparrow$ & Style Sim. $\uparrow$ & Comb. Score $\uparrow$ \\
            \midrule
             &  &  & 0.65 & 0.60 & 0.62 \\
            \checkmark &  &  & 0.73 & 0.70 & 0.72 \\
             & \checkmark &  & 0.69 & 0.65 & 0.67 \\
             &  & \checkmark & 0.68 & 0.64 & 0.66 \\
            \checkmark & \checkmark &  & 0.76 & 0.76 & 0.80 \\
            \checkmark &  & \checkmark & 0.75 & 0.73 & 0.78 \\
             & \checkmark & \checkmark & 0.71 & 0.68 & 0.70 \\
            \rowcolor{gray!15}
            \checkmark & \checkmark & \checkmark & \textbf{0.79} & \textbf{0.80} & \textbf{0.83} \\
            \bottomrule
        \end{tabular}
        }
    \end{subfigure}   
    \begin{subfigure}[t]{1.0\linewidth}
        \centering
        \caption{User Study}
        \label{subfig:user_study}
        \resizebox{\linewidth}{!}{%
        \begin{tabular}{@{}lccc@{}}
            \toprule
            Method & Content Fidelity $\uparrow$ & Style Fidelity $\uparrow$ & Coherence $\uparrow$ \\
            \midrule
            Direct Merging & 2.3 & 2.0 & 2.1 \\
            StyleDrop      & 2.8 & 3.6 & 3.0 \\
            ZipLoRA        & 3.1 & 3.3 & 3.4 \\
            KLoRA          & 3.3 & 3.5 & 3.5 \\
            BLoRA          & 3.5 & 3.4 & 3.6 \\
            \rowcolor{gray!15}
            \textbf{Ours}  & \textbf{4.1} & \textbf{4.3} & \textbf{4.4} \\
            \bottomrule
        \end{tabular}
        }
    \end{subfigure}
    \caption{Quantitative evaluation of our method. 
    (a) Comparison against baselines using CLIP-I similarities and GPT-4o binary judgment scores. 
    (b) Component ablation: Rank-FT contributes most to disentanglement, Router provides content-style routing, and ACFG stabilizes fusion; all three are complementary. The baseline (no components) corresponds to Direct Merging. 
    (c) User study ratings on a 1--5 scale (30 participants, 50 samples).}
    \label{fig:quantitative_data}
\end{figure}

\subsection{Results and Analysis}
In this section, we present a comprehensive analysis of our proposed method, encompassing visual demonstrations, quantitative evaluations against baselines, and an ablation study investigating the contributions of key components.

\textbf{Visual Results.}  
Figure~\ref{fig:visual_results1} compares our method with representative LoRA fusion baselines, including ZipLoRA, BLoRA, and KLoRA. Prompts are formatted as ``A [content] \texttt{<c>} in [style] \texttt{<s>}'' with different content–style pairs such as dogs, plush toys, and objects. While existing methods often compromise either content fidelity \eg structural distortion or style rendering \eg muted patterns, our approach produces outputs that preserve object identity while faithfully capturing artistic styles.  
Figure~\ref{fig:visual_results2} provides extended qualitative demonstrations. The top part shows content–style combinations augmented with additional prompt descriptions \eg ``catching a frisbee'', ``playing with a ball'', or ``driving a car''. Our method integrates these dynamic semantics while faithfully maintaining the designated content and style. The bottom part isolates the effect of single-branch adapters. Activating only the content branch preserves structural fidelity of objects across neutral renderings, while activating only the style branch applies stylistic transformations consistently across varied subjects. These observations confirm that our framework achieves effective disentanglement and controllability during inference.

\textbf{Quantitative Evaluation.} 
We evaluate our method using both automatic and human-centered metrics. For automatic evaluation, we adopt \textit{Content Similarity} and \textit{Style Similarity} based on CLIP-I embeddings, as well as a \textit{Combination Score} derived from GPT-4o that measures the overall coherence of content–style integration. Each score is averaged over 10 distinct content–style pairs, with 10 generated images per pair. 
As shown in Table~\ref{fig:quantitative_data}~(a), baseline methods exhibit different trade-offs: Direct Merging fails to preserve either content or style, StyleDrop emphasizes style at the expense of structure, and ZipLoRA, KLoRA, BLoRA, and LoRA.rar achieve more balanced results but remain limited in overall coherence. Our method consistently achieves the highest scores across all metrics, demonstrating its ability to preserve structural fidelity while maintaining faithful style transfer and coherent integration.

\textbf{Ablation and User Study.} 
Table~\ref{fig:quantitative_data}~(b) presents a comprehensive component ablation. Among individual components, Rank-FT contributes most significantly, improving Content Sim. by +0.08 and Style Sim. by +0.10 over the baseline, confirming its central role in content-style disentanglement. The Router provides complementary gains through layer-wise content-style routing, while ACFG stabilizes fusion quality. Combining any two components yields further improvements, and the full configuration achieves the best performance across all metrics, demonstrating that the three components are complementary.

Table~\ref{fig:quantitative_data}~(c) reports results from a user study with 30 participants rating 50 samples on a 1--5 scale. Our method achieves the highest ratings across content fidelity, style fidelity, and overall coherence, consistent with the automatic metrics. This combination of quantitative and perceptual evidence confirms that our framework achieves superior disentanglement, fidelity, and integration compared with prior approaches.

\textbf{Failure Cases Analysis.}
Despite the effectiveness of our approach, we identified three main failure modes during the diverse implementation process.
First, when the reference image contains highly entangled content and style (\eg a cartoon character whose identity is inseparable from the rendering), the method struggles to isolate content from style, resulting in style leakage into the content branch. 
Second, for styles dominated by low-frequency characteristics such as flat color palettes, the frequency-based separation misassigns these features to the content domain, leading to incomplete style capture. 
Third, abstract style descriptions (\eg ``ethereal mood'') may lack precise CLIP embeddings, resulting in weak or inconsistent style transfer. Adjusting the frequency cutoff $\sigma$ or adopting stronger encoders could mitigate these issues.
\section{Conclusion}
\label{sec:conclusion}
We presented a framework for personalized image generation that integrates efficient adaptation strategies. Rank-constrained backbone fine-tuning encourages distinct subspaces, providing an inductive bias for disentanglement. An expert encoder with selective aggregation enhances semantic guidance, enabling flexible composition, and timestep-dependent asymmetric CFG introduces a training-free sampling strategy that improves visual quality. Our experiments demonstrate measurable reduction in content-style cross-influence, consistent improvements across automatic metrics, and clear user preference in perceptual studies. 
Limitations include architecture-dependent layer selection, reliance on text embedding quality, frequency-based separation assumptions that may not hold for flat textures, and the two-branch ACFG structure which limits multi-content or multi-style mixing. Standard SDXL LoRAs can be used with ACFG directly on the base model, though full disentanglement benefits require retraining on the fine-tuned backbone. 
Future work will explore automated layer assignment, multi-concept scheduling, and extensions to architectures such as KOALA~\cite{lee2024koala} and SANA~\cite{xie2024sana}.
\clearpage

\section*{Acknowledgement}
This work was supported by the National Natural Science Foundation of China (No. 6250070674) and the Zhejiang Leading Innovative and Entrepreneur Team Introduction Program (2024R01007).

{
    \small
    \bibliographystyle{ieeenat}
    \bibliography{ref}

@article{podell2023sdxl,
  title={Sdxl: Improving latent diffusion models for high-resolution image synthesis},
  author={Podell, Dustin and English, Zion and Lacey, Kyle and Blattmann, Andreas and Dockhorn, Tim and M{\"u}ller, Jonas and Penna, Joe and Rombach, Robin},
  journal={arXiv preprint arXiv:2307.01952},
  year={2023}
}

@inproceedings{ruiz2023dreambooth,
  title={Dreambooth: Fine tuning text-to-image diffusion models for subject-driven generation},
  author={Ruiz, Nataniel and Li, Yuanzhen and Jampani, Varun and Pritch, Yael and Rubinstein, Michael and Aberman, Kfir},
  booktitle={Proceedings of the IEEE/CVF conference on computer vision and pattern recognition},
  pages={22500--22510},
  year={2023}
}

@article{sohn2023styledrop,
  title={Styledrop: Text-to-image synthesis of any style},
  author={Sohn, Kihyuk and Jiang, Lu and Barber, Jarred and Lee, Kimin and Ruiz, Nataniel and Krishnan, Dilip and Chang, Huiwen and Li, Yuanzhen and Essa, Irfan and Rubinstein, Michael and others},
  journal={Advances in Neural Information Processing Systems},
  volume={36},
  pages={66860--66889},
  year={2023}
}

@article{cai2021generative,
  title={Generative adversarial networks: A survey toward private and secure applications},
  author={Cai, Zhipeng and Xiong, Zuobin and Xu, Honghui and Wang, Peng and Li, Wei and Pan, Yi},
  journal={ACM Computing Surveys (CSUR)},
  volume={54},
  number={6},
  pages={1--38},
  year={2021},
  publisher={ACM New York, NY, USA}
}

@inproceedings{rombach2022high,
  title={High-resolution image synthesis with latent diffusion models},
  author={Rombach, Robin and Blattmann, Andreas and Lorenz, Dominik and Esser, Patrick and Ommer, Bj{\"o}rn},
  booktitle={Proceedings of the IEEE/CVF conference on computer vision and pattern recognition},
  pages={10684--10695},
  year={2022}
}

@inproceedings{blattmann2023align,
  title={Align your latents: High-resolution video synthesis with latent diffusion models},
  author={Blattmann, Andreas and Rombach, Robin and Ling, Huan and Dockhorn, Tim and Kim, Seung Wook and Fidler, Sanja and Kreis, Karsten},
  booktitle={Proceedings of the IEEE/CVF conference on computer vision and pattern recognition},
  pages={22563--22575},
  year={2023}
}

@article{hertz2022prompt,
  title={Prompt-to-prompt image editing with cross attention control},
  author={Hertz, Amir and Mokady, Ron and Tenenbaum, Jay and Aberman, Kfir and Pritch, Yael and Cohen-Or, Daniel},
  journal={arXiv preprint arXiv:2208.01626},
  year={2022}
}

@article{hao2023vico,
  title={Vico: Plug-and-play visual condition for personalized text-to-image generation},
  author={Hao, Shaozhe and Han, Kai and Zhao, Shihao and Wong, Kwan-Yee K},
  journal={arXiv preprint arXiv:2306.00971},
  year={2023}
}

@inproceedings{zhang2024pia,
  title={Pia: Your personalized image animator via plug-and-play modules in text-to-image models},
  author={Zhang, Yiming and Xing, Zhening and Zeng, Yanhong and Fang, Youqing and Chen, Kai},
  booktitle={Proceedings of the IEEE/CVF conference on computer vision and pattern recognition},
  pages={7747--7756},
  year={2024}
}

@inproceedings{chung2024style,
  title={Style injection in diffusion: A training-free approach for adapting large-scale diffusion models for style transfer},
  author={Chung, Jiwoo and Hyun, Sangeek and Heo, Jae-Pil},
  booktitle={Proceedings of the IEEE/CVF conference on computer vision and pattern recognition},
  pages={8795--8805},
  year={2024}
}

@article{li2024beyond,
  title={Beyond inserting: Learning identity embedding for semantic-fidelity personalized diffusion generation},
  author={Li, Yang and Yang, Songlin and Wang, Wei and Dong, Jing},
  journal={arXiv preprint arXiv:2402.00631},
  year={2024}
}

@article{shuai2024survey,
  title={A survey of multimodal-guided image editing with text-to-image diffusion models},
  author={Shuai, Xincheng and Ding, Henghui and Ma, Xingjun and Tu, Rongcheng and Jiang, Yu-Gang and Tao, Dacheng},
  journal={arXiv preprint arXiv:2406.14555},
  year={2024}
}

@inproceedings{tanveer2023ds,
  title={Ds-fusion: Artistic typography via discriminated and stylized diffusion},
  author={Tanveer, Maham and Wang, Yizhi and Mahdavi-Amiri, Ali and Zhang, Hao},
  booktitle={Proceedings of the IEEE/CVF International Conference on Computer Vision},
  pages={374--384},
  year={2023}
}

@article{park2023visual,
  title={Visual language integration: A survey and open challenges},
  author={Park, Sang-Min and Kim, Young-Gab},
  journal={Computer Science Review},
  volume={48},
  pages={100548},
  year={2023},
  publisher={Elsevier}
}

@inproceedings{ruiz2024hyperdreambooth,
  title={Hyperdreambooth: Hypernetworks for fast personalization of text-to-image models},
  author={Ruiz, Nataniel and Li, Yuanzhen and Jampani, Varun and Wei, Wei and Hou, Tingbo and Pritch, Yael and Wadhwa, Neal and Rubinstein, Michael and Aberman, Kfir},
  booktitle={Proceedings of the IEEE/CVF conference on computer vision and pattern recognition},
  pages={6527--6536},
  year={2024}
}

@article{gal2022image,
  title={An image is worth one word: Personalizing text-to-image generation using textual inversion},
  author={Gal, Rinon and Alaluf, Yuval and Atzmon, Yuval and Patashnik, Or and Bermano, Amit H and Chechik, Gal and Cohen-Or, Daniel},
  journal={arXiv preprint arXiv:2208.01618},
  year={2022}
}

@article{tang2023parameter,
  title={Parameter efficient multi-task model fusion with partial linearization},
  author={Tang, Anke and Shen, Li and Luo, Yong and Zhan, Yibing and Hu, Han and Du, Bo and Chen, Yixin and Tao, Dacheng},
  journal={arXiv preprint arXiv:2310.04742},
  year={2023}
}

@inproceedings{shah2024ziplora,
  title={Ziplora: Any subject in any style by effectively merging loras},
  author={Shah, Viraj and Ruiz, Nataniel and Cole, Forrester and Lu, Erika and Lazebnik, Svetlana and Li, Yuanzhen and Jampani, Varun},
  booktitle={European Conference on Computer Vision},
  pages={422--438},
  year={2024},
  organization={Springer}
}

@inproceedings{frenkel2024implicit,
  title={Implicit style-content separation using b-lora},
  author={Frenkel, Yarden and Vinker, Yael and Shamir, Ariel and Cohen-Or, Daniel},
  booktitle={European Conference on Computer Vision},
  pages={181--198},
  year={2024},
  organization={Springer}
}

@article{zhong2024multi,
  title={Multi-lora composition for image generation},
  author={Zhong, Ming and Shen, Yelong and Wang, Shuohang and Lu, Yadong and Jiao, Yizhu and Ouyang, Siru and Yu, Donghan and Han, Jiawei and Chen, Weizhu},
  journal={arXiv preprint arXiv:2402.16843},
  year={2024}
}

@article{wang2024lora,
  title={Lora-flow: Dynamic lora fusion for large language models in generative tasks},
  author={Wang, Hanqing and Ping, Bowen and Wang, Shuo and Han, Xu and Chen, Yun and Liu, Zhiyuan and Sun, Maosong},
  journal={arXiv preprint arXiv:2402.11455},
  year={2024}
}

@article{ghebrehiwet2024revolutionizing,
  title={Revolutionizing personalized medicine with generative AI: a systematic review},
  author={Ghebrehiwet, Isaias and Zaki, Nazar and Damseh, Rafat and Mohamad, Mohd Saberi},
  journal={Artificial Intelligence Review},
  volume={57},
  number={5},
  pages={128},
  year={2024},
  publisher={Springer}
}

@article{bengesi2024advancements,
  title={Advancements in Generative AI: A Comprehensive Review of GANs, GPT, Autoencoders, Diffusion Model, and Transformers.},
  author={Bengesi, Staphord and El-Sayed, Hoda and Sarker, Md Kamruzzaman and Houkpati, Yao and Irungu, John and Oladunni, Timothy},
  journal={IEEe Access},
  year={2024},
  publisher={IEEE}
}

@article{sengar2024generative,
  title={Generative artificial intelligence: a systematic review and applications},
  author={Sengar, Sandeep Singh and Hasan, Affan Bin and Kumar, Sanjay and Carroll, Fiona},
  journal={Multimedia Tools and Applications},
  pages={1--40},
  year={2024},
  publisher={Springer}
}

@article{hu2022lora,
  title={Lora: Low-rank adaptation of large language models.},
  author={Hu, Edward J and Shen, Yelong and Wallis, Phillip and Allen-Zhu, Zeyuan and Li, Yuanzhi and Wang, Shean and Wang, Lu and Chen, Weizhu and others},
  journal={ICLR},
  volume={1},
  number={2},
  pages={3},
  year={2022}
}

@article{ryu2023low,
  title={Low-rank adaptation for fast text-to-image diffusion fine-tuning},
  author={Ryu, Simo},
  journal={Low-rank adaptation for fast text-to-image diffusion fine-tuning},
  volume={3},
  year={2023}
}

@article{ouyang2025k,
  title={K-lora: Unlocking training-free fusion of any subject and style loras},
  author={Ouyang, Ziheng and Li, Zhen and Hou, Qibin},
  journal={arXiv preprint arXiv:2502.18461},
  year={2025}
}

@article{liu2022learning,
  title={Learning disentangled representations in the imaging domain},
  author={Liu, Xiao and Sanchez, Pedro and Thermos, Spyridon and O’Neil, Alison Q and Tsaftaris, Sotirios A},
  journal={Medical Image Analysis},
  volume={80},
  pages={102516},
  year={2022},
  publisher={Elsevier}
}

@article{liu2024unziplora,
  title={UnZipLoRA: Separating Content and Style from a Single Image},
  author={Liu, Chang and Shah, Viraj and Cui, Aiyu and Lazebnik, Svetlana},
  journal={arXiv preprint arXiv:2412.04465},
  year={2024}
}

@inproceedings{pavllo2020controlling,
  title={Controlling style and semantics in weakly-supervised image generation},
  author={Pavllo, Dario and Lucchi, Aurelien and Hofmann, Thomas},
  booktitle={Computer Vision--ECCV 2020: 16th European Conference, Glasgow, UK, August 23--28, 2020, Proceedings, Part VI 16},
  pages={482--499},
  year={2020},
  organization={Springer}
}

@inproceedings{liu2023more,
  title={More control for free! image synthesis with semantic diffusion guidance},
  author={Liu, Xihui and Park, Dong Huk and Azadi, Samaneh and Zhang, Gong and Chopikyan, Arman and Hu, Yuxiao and Shi, Humphrey and Rohrbach, Anna and Darrell, Trevor},
  booktitle={Proceedings of the IEEE/CVF winter conference on applications of computer vision},
  pages={289--299},
  year={2023}
}

@article{wang2024parameter,
  title={Parameter-efficient fine-tuning in large models: A survey of methodologies},
  author={Wang, Luping and Chen, Sheng and Jiang, Linnan and Pan, Shu and Cai, Runze and Yang, Sen and Yang, Fei},
  journal={arXiv preprint arXiv:2410.19878},
  year={2024}
}

@article{han2024parameter,
  title={Parameter-efficient fine-tuning for large models: A comprehensive survey},
  author={Han, Zeyu and Gao, Chao and Liu, Jinyang and Zhang, Jeff and Zhang, Sai Qian},
  journal={arXiv preprint arXiv:2403.14608},
  year={2024}
}

@inproceedings{avrahami2023break,
  title={Break-a-scene: Extracting multiple concepts from a single image},
  author={Avrahami, Omri and Aberman, Kfir and Fried, Ohad and Cohen-Or, Daniel and Lischinski, Dani},
  booktitle={SIGGRAPH Asia 2023 Conference Papers},
  pages={1--12},
  year={2023}
}

@article{chen2024personalizing,
  title={PaRa: Personalizing Text-to-Image Diffusion via Parameter Rank Reduction},
  author={Chen, Shangyu and Pan, Zizheng and Cai, Jianfei and Phung, Dinh},
  journal={arXiv preprint arXiv:2406.05641},
  year={2024}
}

@inproceedings{finn2017model,
  title={Model-agnostic meta-learning for fast adaptation of deep networks},
  author={Finn, Chelsea and Abbeel, Pieter and Levine, Sergey},
  booktitle={International conference on machine learning},
  pages={1126--1135},
  year={2017},
  organization={PMLR}
}

@article{chua2021fine,
  title={How fine-tuning allows for effective meta-learning},
  author={Chua, Kurtland and Lei, Qi and Lee, Jason D},
  journal={Advances in Neural Information Processing Systems},
  volume={34},
  pages={8871--8884},
  year={2021}
}

@article{zhao2022effectiveness,
  title={On the effectiveness of fine-tuning versus meta-reinforcement learning},
  author={Zhao, Mandi and Abbeel, Pieter and James, Stephen},
  journal={Advances in neural information processing systems},
  volume={35},
  pages={26519--26531},
  year={2022}
}

@article{li2025dual,
  title={Dual branch segment anything model-transformer fusion network for accurate breast ultrasound image segmentation},
  author={Li, Yu and Huang, Jin and Zhang, Yimin and Deng, Jingwen and Zhang, Jingwen and Dong, Lan and Wang, Du and Mei, Liye and Lei, Cheng},
  journal={Medical Physics},
  volume={52},
  number={6},
  pages={4108--4119},
  year={2025},
  publisher={Wiley Online Library}
}

@inproceedings{ma2025scflow,
  title={SCFlow: Implicitly Learning Style and Content Disentanglement with Flow Models},
  author={Ma, Pingchuan and Yang, Xiaopei and Li, Yusong and Gui, Ming and Krause, Felix and Schusterbauer, Johannes and Ommer, Bj{\"o}rn},
  booktitle={Proceedings of the IEEE/CVF International Conference on Computer Vision},
  pages={14919--14929},
  year={2025}
}

@inproceedings{yang2025qr,
  title={Qr-lora: Efficient and disentangled fine-tuning via qr decomposition for customized generation},
  author={Yang, Jiahui and Ma, Yongjia and Di, Donglin and Cui, Jianxun and Li, Hao and Chen, Wei and Xie, Yan and Yang, Xun and Zuo, Wangmeng},
  booktitle={Proceedings of the IEEE/CVF International Conference on Computer Vision},
  pages={17587--17597},
  year={2025}
}

@inproceedings{shenaj2025lora,
  title={Lora. rar: Learning to merge loras via hypernetworks for subject-style conditioned image generation},
  author={Shenaj, Donald and Bohdal, Ondrej and Ozay, Mete and Zanuttigh, Pietro and Michieli, Umberto},
  booktitle={Proceedings of the IEEE/CVF International Conference on Computer Vision},
  pages={16132--16142},
  year={2025}
}

@inproceedings{meral2025contrastive,
  title={Contrastive test-time composition of multiple LoRA models for image generation},
  author={Meral, Tuna Han Salih and Simsar, Enis and Tombari, Federico and Yanardag, Pinar},
  booktitle={Proceedings of the IEEE/CVF International Conference on Computer Vision},
  pages={18090--18100},
  year={2025}
}

@article{lee2024koala,
  title={Koala: Empirical lessons toward memory-efficient and fast diffusion models for text-to-image synthesis},
  author={Lee, Youngwan and Park, Kwanyong and Cho, Yoorhim and Lee, Yong-Ju and Hwang, Sung Ju},
  journal={Advances in Neural Information Processing Systems},
  volume={37},
  pages={51597--51633},
  year={2024}
}

@article{xie2024sana,
  title={Sana: Efficient high-resolution image synthesis with linear diffusion transformers},
  author={Xie, Enze and Chen, Junsong and Chen, Junyu and Cai, Han and Tang, Haotian and Lin, Yujun and Zhang, Zhekai and Li, Muyang and Zhu, Ligeng and Lu, Yao and others},
  journal={arXiv preprint arXiv:2410.10629},
  year={2024}
}

@inproceedings{lin2024evaluating,
  title={Evaluating text-to-visual generation with image-to-text generation},
  author={Lin, Zhiqiu and Pathak, Deepak and Li, Baiqi and Li, Jiayao and Xia, Xide and Neubig, Graham and Zhang, Pengchuan and Ramanan, Deva},
  booktitle={European Conference on Computer Vision},
  pages={366--384},
  year={2024},
  organization={Springer}
}
}


\clearpage
\setcounter{page}{1}
\maketitlesupplementary

\appendix
\section*{Appendix}
\label{sec:appendix}


\section{Trunk (Backbone) Fine-tuning Details}

\subsection{Targeted Dataset Construction}
We construct a dataset of 100 content–style contrast pairs, \(\mathcal{D}=\{(I_c^k, I_s^k)\}_{k=1}^{100}\). This dataset is used to fine-tune the backbone, encouraging content–style disentanglement and providing the base weights (\(W_{\text{init}}\)) for subsequent LoRA adapter training. The pairs are generated using a pre-trained diffusion model with frequency-based controls.

For each pair \((I_c^k, I_s^k)\in\mathcal{D}\): image \(I_c^k\) emphasizes content, guided by a content prompt \(P_c^k\) and a style modifier \(P_{sm}^k\); image \(I_s^k\) emphasizes style, guided by a style prompt \(P_s^k\) and a content modifier \(P_{cm}^k\). One specific \(P_{sm}^k\) and one \(P_{cm}^k\) are chosen per pair to form a one-to-one contrast.

We employ frequency-domain manipulations during iterative denoising. Let \(\mathcal{F}\) denote a frequency transform (e.g., Discrete Cosine Transform) and \(\mathcal{F}^{-1}\) its inverse. Masks \(\mathcal{M}_{\text{low}}\) and \(\mathcal{M}_{\text{high}}\) preserve low- and high-frequency components, respectively. With noisy latent \(z_t\) at timestep \(t\) and text embedding \(e=\mathrm{TextualEmbeddings}(P)\), the process is:
\begin{align*}
\hat{z}_0 &= \mathrm{Predictor}(z_t,t,e), \quad \text{(Predict clean latent)} \\
z_{t-1}^{\text{filtered}} &= \mathcal{F}^{-1}\!\big(\mathcal{M}_{\text{low}}\odot \mathcal{F}(\hat{z}_0)\big)\quad \text{(for } I_c^k \text{ generation)},\\
z_{t-1}^{\text{filtered}} &= \mathcal{F}^{-1}\!\big(\mathcal{M}_{\text{high}}\odot \mathcal{F}(\hat{z}_0)\big)\quad \text{(for } I_s^k \text{ generation)},\\
z_{t-1} &= \mathrm{UpdateRule}(z_t, z_{t-1}^{\text{filtered}}, \dots),
\end{align*}
where \(\odot\) denotes element-wise multiplication. This design concentrates variation along the intended factor (content or style).

\begin{table}[htbp!]
\centering
\caption{Conceptual illustration of content–style contrast pair generation. For each pair \((I_c^k, I_s^k)\), one varying modifier is chosen for \(I_c^k\) and one for \(I_s^k\).}
\begingroup
\setlength{\tabcolsep}{6pt}
\renewcommand{\arraystretch}{1.15}
\resizebox{\linewidth}{!}{
\begin{tabular}{@{}p{0.42\linewidth}cc@{}}
\toprule
\textbf{Property} & \textbf{\(I_c^k\) (content fixed)} & \textbf{\(I_s^k\) (style fixed)} \\
\midrule
Target image & Generated with fixed content & Generated with fixed style \\
Base prompt & \(P_c^k\): ``a red car'' & \(P_s^k\): ``in the style of Van Gogh'' \\
Varying modifier & \(P_{sm}^k\in\{\text{watercolor},\ \text{oil painting},\dots\}\) & \(P_{cm}^k\in\{\text{starry night},\ \text{sunflower},\dots\}\) \\
Dominant frequency filter & Low \(\mathcal{M}_{\text{low}}\) & High \(\mathcal{M}_{\text{high}}\) \\
Resulting characteristic & Content preserved with varied textures & Style preserved with varied subjects \\
\bottomrule
\end{tabular}}
\endgroup
\label{tab:contrast_pair_generation}
\end{table}

\subsection{Training Settings}
We fine-tune the backbone weights \(\{W_l\}_{l=1}^L\) by minimizing:
\begin{align}
\mathcal L_{\text{trunk}}=\mathcal L_{\text{task}}(\{W_l\},\mathcal{D})+\lambda_{\text{reg}}\sum_{l=1}^L \|B_l\|_F^2,
\label{eq:trunk_loss}
\end{align}
where \(B_l\) is a learnable basis for each layer \(l\). Let \(B_l=Q_lR_l\) be the QR decomposition of the basis, where \(Q_l\) contains the orthonormal vectors (\(Q_l^\top Q_l=I\)). Consistent with the update rule discussed in the main text, the weights are updated as:
\begin{equation}
W_l = W_l^{(0)} - Q_l Q_l^\top W_l^{(0)}.
\label{eq:trunk_reduce}
\end{equation}
This update effectively projects the original weights \(W_l^{(0)}\) onto the orthogonal complement of the learned subspace \(\mathrm{span}(Q_l)\). We denote the resulting post-finetuning backbone as \(W_{\text{init}}\) (also written \(\widetilde{W}\) in figures). The update \(\Delta W_l=-Q_l Q_l^\top W_l^{(0)}\) lies in \(\mathrm{span}(Q_l)\).

The per-layer rank \(r_l\) for the basis \(B_l\) follows a linear schedule:
\begin{align}
r_l = r_{\max} - \frac{l-1}{L-1}\big(r_{\max}-r_{\min}\big), \quad l=1,\dots,L,
\label{eq:rank_schedule}
\end{align}
with maximum rank \(r_{\max}=128\) and minimum rank \(r_{\min}=4\).


\begin{figure*}[t] 
\begin{multline}
\mathcal L_{\text{task}}(\{W_l\}, \mathcal{D})
= \frac{1}{N} \sum_{k=1}^{N} \mathbb{E}_{t, z_t} \Bigg[
\big\|G(\{W_l\}, z_t, t, \mathrm{TextualEmbeddings}(P_c^k, P_{sm}^k)) - I_c^k \big\|_1 \\
+ \big\|G(\{W_l\}, z_t, t, \mathrm{TextualEmbeddings}(P_{cm}^k, P_s^k)) - I_s^k \big\|_1 \Bigg]
+ \alpha\, \mathcal L_{\text{perceptual}}(\{W_l\}, \mathcal{D}).
\label{eq:task_loss}
\end{multline}
\end{figure*}

\begin{figure*}[t] 
\begin{multline}
\mathcal L_{\text{perceptual}}
= \frac{1}{N}\sum_{k=1}^{N}\mathbb{E}_{t, z_t}\sum_{j\in\mathrm{VGG\_layers}}\frac{1}{M_j}\Bigg[
\big\|\phi_j\big(G(\{W_l\}, z_t, t, \mathrm{TextualEmbeddings}(P_c^k, P_{sm}^k))\big)-\phi_j(I_c^k)\big\|_1 \\
+ \big\|\phi_j\big(G(\{W_l\}, z_t, t, \mathrm{TextualEmbeddings}(P_{cm}^k, P_s^k))\big)-\phi_j(I_s^k)\big\|_1 \Bigg].
\label{eq:perceptual_loss}
\end{multline}
\end{figure*}

The task loss \(\mathcal L_{\text{task}}\) (Eq.~\ref{eq:task_loss}) is computed over the dataset \(\mathcal{D}\).
Here, \(N=100\), \(G(\cdot)\) represents the \(\hat{x}_0\) predictor (i.e., the U-Net), \(t\) is a sampled timestep, and \(z_t\) is the corresponding noisy latent created from the target image.

The perceptual term \(\mathcal L_{\text{perceptual}}\) (Eq.~\ref{eq:perceptual_loss}) is defined as:
where \(\phi_j\) are features from VGG-19 (relu1\_1 to relu5\_1) and \(M_j\) is the number of elements in the feature map. We set \(\alpha=0.1\) and \(\lambda_{\text{reg}}=\num{1e-4}\).

Optimization uses the AdamW optimizer with a cosine decay learning rate schedule and a 500-step warm-up. The learning rate starts at \(\num{1e-5}\), warms up from \(\num{1e-6}\), and decays to a minimum of \(\num{1e-7}\). We use a batch size of 8 and train for 5000 steps.

\begin{table}[htbp!]
\centering
\caption{Summary of backbone fine-tuning hyperparameters.}
\begingroup
\setlength{\tabcolsep}{4pt}
\renewcommand{\arraystretch}{1.1}
\resizebox{\linewidth}{!}{
\begin{tabular}{@{}ll@{}}
\toprule
Parameter & Value \\
\midrule
Optimizer & AdamW \\
Initial learning rate & \num{1e-5} \\
LR schedule & Cosine decay with 500 warm-up steps \\
Minimum learning rate & \num{1e-7} \\
Batch size & 8 \\
Optimization steps & 5000 \\
Tuned U-Net layers & Attention and standard convolutional layers \\
Rank schedule \(r_l\) (Eq.~\ref{eq:rank_schedule}) & Linear; \(r_{\max}=128\), \(r_{\min}=4\) \\
\(\mathcal L_{\text{task}}\) components & \(\ell_1\) + Perceptual (VGG-19 relu1\_1 to relu5\_1) \\
Perceptual loss weight \(\alpha\) & 0.1 \\
\(B_l\) regularization \(\lambda_{\text{reg}}\) & \num{1e-4} \\
Hardware & 2 NVIDIA 4090 GPUs \\
\bottomrule
\end{tabular}}
\endgroup
\label{tab:training_settings_detailed}
\end{table}

\section{LoRA Training and Inference Details}
\label{app:lora}

\subsection{Setting and Notation}
LoRA adapters are trained on the U-Net backbone initialized with the pre-tuned weights \(W_{\text{init}}\) (from Eq.~\ref{eq:trunk_reduce}). We define disjoint sets of U-Net layers \(I_c\) (for content) and \(I_s\) (for style), satisfying \(I_c\cap I_s=\emptyset\). The adapters are trained only on the layers specified in \(I_c\cup I_s\).

\textbf{LoRA parameterization.} For any layer \(i\in I_c\cup I_s\), we use a fixed LoRA rank \(r=16\). Content adapters learn parameters \(\{B_i^{(c)},A_i^{(c)}\}\) for \(i\in I_c\); style adapters learn parameters \(\{B_i^{(s)},A_i^{(s)}\}\) for \(i\in I_s\). The update is \(\Delta W_i = B_i A_i\).

\textbf{Training procedure.} During content adapter training, gradients for \(\{B_i^{(c)},A_i^{(c)}\}\) are enabled only for \(i\in I_c\) (and masked for \(i\in I_s\)). Conversely, style adapter training only updates parameters for \(i\in I_s\). Training hyperparameters are listed in Table~\ref{tab:lora_training_hyperparams}. The specific layer assignments for content and style are shown in Table~\ref{tab:layer_split_lora}.

\begin{table}[htbp!]
\centering
\caption{LoRA adapter training hyperparameters.}
\begingroup
\begin{tabular}{@{}ll@{}}
\toprule
Parameter & Value \\
\midrule
Optimizer & AdamW \\
Learning rate & \num{1e-5} \\
Training steps & 1000–2000 \\
Batch size & 1 \\
Backbone host & \(W_{\text{init}}\) (from Eq.~\ref{eq:trunk_reduce}) \\
\bottomrule
\end{tabular}
\endgroup
\label{tab:lora_training_hyperparams}
\end{table}

\begin{table*}[htbp!]
\centering
\caption{U-Net layer assignments for content \((I_c)\) and style \((I_s)\) adapters.}
\begingroup
\setlength{\tabcolsep}{5pt}
\renewcommand{\arraystretch}{1.12}
\resizebox{\linewidth}{!}{
\begin{tabular}{@{}p{0.47\linewidth}p{0.47\linewidth}@{}}
\toprule
\textbf{Content layers \((I_c)\)} & \textbf{Style layers \((I_s)\)} \\
\midrule
\texttt{down\_blocks.2.attentions.0}; \texttt{mid\_block.attentions.0}; \texttt{up\_blocks.0.attentions.1}; \texttt{up\_blocks.1.attentions.0}; \texttt{up\_blocks.2.attentions.0}
&
\texttt{down\_blocks.0.attentions.0}; \texttt{down\_blocks.1.resnets.0}; \texttt{mid\_block.resnets.1}; \texttt{down\_blocks.0.resnets.1}; \texttt{down\_blocks.1.attentions.1} \\
\bottomrule
\end{tabular}}
\endgroup
\label{tab:layer_split_lora}
\end{table*}

\textbf{Expert encoder and aggregation.} Three MLP encoders, \(E_t,E_c,E_s\), produce 64-dimensional embeddings from a concept ID embedding, a CLIP content text embedding, and a CLIP style text embedding, respectively. Their concatenated features are fed into an aggregation head that outputs nonnegative scaling factors \((\gamma_c,\gamma_s)\). At inference, the aggregated weights are:
\[
W^{\mathrm{agg}}
= W_{\text{init}}
+ \sum_{i\in I_c}\mathsf{E}_i\!\big(\gamma_c\,\Delta W^{(c)}_i\big)
+ \sum_{i\in I_s}\mathsf{E}_i\!\big(\gamma_s\,\Delta W^{(s)}_i\big),
\]
where \(\mathsf{E}_i(\cdot)\) is an operator that injects the low-rank update \(\Delta W_i = B_i A_i\) at layer \(i\).

\subsection{Disentanglement Metrics and Results}
\begin{table}[htbp!]
\centering
\caption{Disentanglement evaluation metrics. \(\mathcal{G}(p_c, p_s)\) denotes the final image generated from content prompt \(p_c\) and style prompt \(p_s\). Higher is better for \(\mathcal{S}_c,\mathcal{S}_s\); lower is better for \(\mathcal{S}_x\).}
\begingroup
\setlength{\tabcolsep}{6pt}
\renewcommand{\arraystretch}{1.12}
\resizebox{\linewidth}{!}{
\begin{tabular}{@{}p{0.26\linewidth}p{0.52\linewidth}p{0.18\linewidth}@{}}
\toprule
\textbf{Metric} & \textbf{Formula} & \textbf{Description} \\
\midrule
Content preservation \(\mathcal{S}_c\) & \(\frac{1}{N_s}\sum_{j=1}^{N_s}\mathrm{sim}_c(\mathcal{G}(p'_c,p_s^j),p'_c)\) & Similarity to target content \\
Style fidelity \(\mathcal{S}_s\) & \(\frac{1}{N_c}\sum_{i=1}^{N_c}\mathrm{sim}_s(\mathcal{G}(p_c^i,p'_s),p'_s)\) & Similarity to target style \\
Cross-influence \(\mathcal{S}_x\) & \(\frac{1}{N_cN_s}\sum_{i,j}\mathrm{interference}(p_c^i,p_s^j)\) & Undesired cross effect \\
\bottomrule
\end{tabular}}
\endgroup
\label{tab:evaluation_metrics}
\end{table}

\begin{table}[htbp!]
\centering
\caption{Quantitative comparison of content–style disentanglement.}
\begingroup
\setlength{\tabcolsep}{8pt}
\renewcommand{\arraystretch}{1.08}
\resizebox{\linewidth}{!}{
\begin{tabular}{@{}lccc@{}}
\toprule
Method & \(\mathcal{S}_c\uparrow\) & \(\mathcal{S}_s\uparrow\) & \(\mathcal{S}_x\downarrow\) \\
\midrule
Standard LoRA & 0.75 & 0.70 & 0.42 \\
Trunk fine-tune + layer-selective LoRA (Ours) & \textbf{0.90} & \textbf{0.88} & \textbf{0.18} \\
\bottomrule
\end{tabular}}
\endgroup
\label{tab:disentanglement_metrics}
\end{table}

\subsection{Semantic Extension Examples}

\begin{table}[htbp!]
\centering
\caption{Types of semantic extension (generalization).}
\begingroup
\setlength{\tabcolsep}{8pt}
\renewcommand{\arraystretch}{1.1}
\resizebox{\linewidth}{!}{
\begin{tabular}{@{}lll@{}}
\toprule
\textbf{Type} & \textbf{Condition} & \textbf{Description} \\
\midrule
I & \(p'_c \neq p_c^i,\ p'_s = p_s^j\) & New content, trained style \\
II & \(p'_c = p_c^i,\ p'_s \neq p_s^j\) & Trained content, new style \\
III & \(p'_c \neq p_c^i,\ p'_s \neq p_s^j\) & New content, new style \\
\bottomrule
\end{tabular}}
\endgroup
\label{tab:extension_types}
\end{table}

\begin{table}[htbp!]
\centering
\caption{Semantic extension examples via prompt variation.}
\begingroup
\setlength{\tabcolsep}{6pt}
\renewcommand{\arraystretch}{1.12}
\resizebox{\linewidth}{!}{
\begin{tabular}{@{}cccc@{}}
\toprule
\textbf{Trained content} & \textbf{Trained style} & \textbf{Inference prompts} & \textbf{Type} \\
\midrule
``a red car'' & ``Van Gogh style'' & \(p'_c\): ``a blue car'', \(p'_s\): ``Van Gogh style'' & I \\
``a red car'' & ``Van Gogh style'' & \(p'_c\): ``a red car'', \(p'_s\): ``sketch style'' & II \\
``a red car'' & ``Van Gogh style'' & \(p'_c\): ``a bicycle'', \(p'_s\): ``sketch style'' & III \\
``person ID X'' & ``pixel art'' & \(p'_c\): ``person ID Y'', \(p'_s\): ``pixel art'' & I \\
``person ID X'' & ``pixel art'' & \(p'_c\): ``person ID X'', \(p'_s\): ``cyberpunk style'' & II \\
\bottomrule
\end{tabular}}
\endgroup
\label{tab:semantic_extension_examples}
\end{table}

\begin{figure*}[htbp!]
    \centering
    \includegraphics[width=\textwidth]{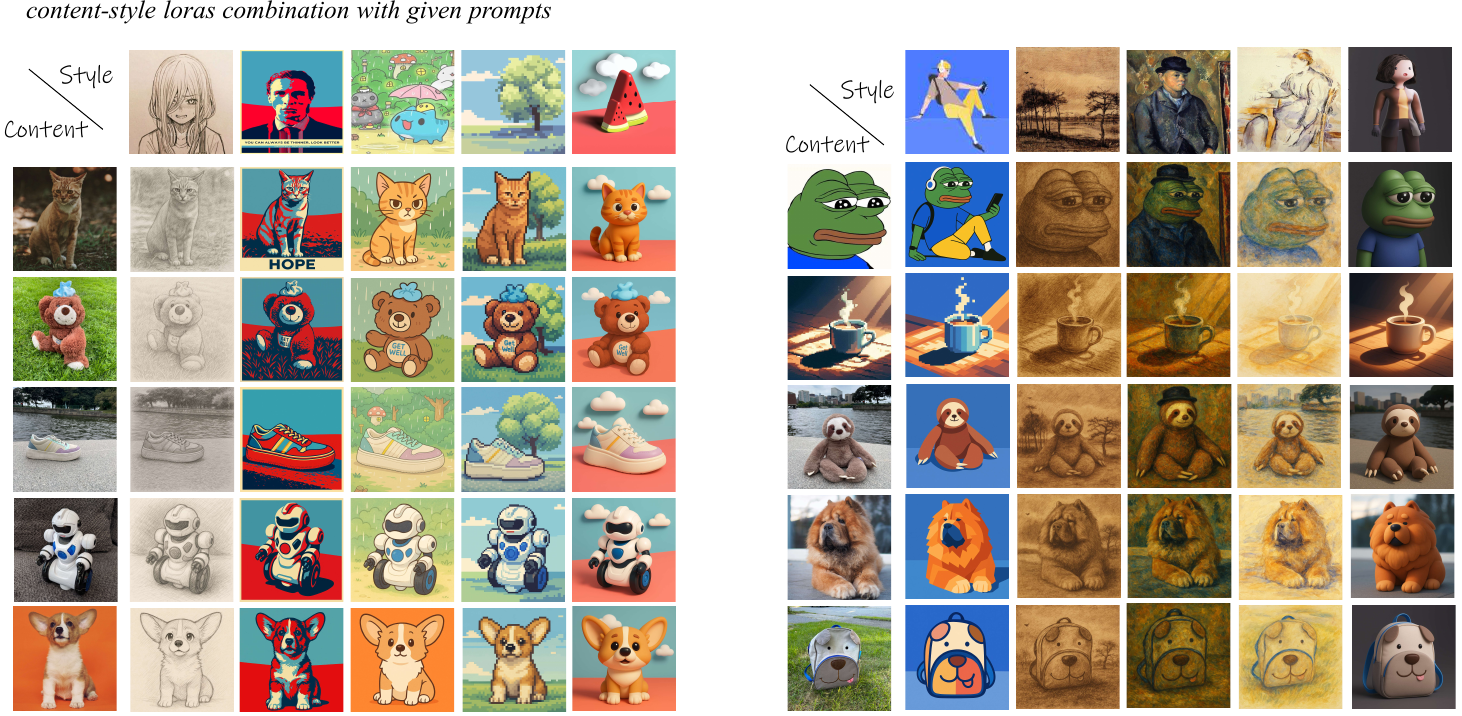}
    \caption{Visual results demonstrating content and style adapter performance.}
    \label{fig:visual_app1}
\end{figure*}

\section{Asymmetric Classifier-Free Guidance (ACFG) for Timestep-Dependent Aggregation}
\label{app:acfg}
We introduce Asymmetric Classifier-Free Guidance (ACFG) during the sampling process. In ACFG, the conditional branch (guidance-seeking) uses the dynamically aggregated LoRA weights, while the unconditional branch (guidance-free) always uses the static, pre-tuned backbone initialization \(W_{\text{init}}\).

\paragraph{Conditional weights.} For timestep \(t\in\{1,\dots,T\}\), the weights for layer \(i\) are:
\begin{equation}
W_i^{\mathrm{cond}}(t)=W_{\text{init},i}
+\gamma_c(t)\,\Delta W^{(c)}_i
+\gamma_s(t)\,\Delta W^{(s)}_i,
\label{eq:cond_weights}
\end{equation}
where \(\gamma_c(t),\gamma_s(t)\ge 0\) are dynamic schedules (which can be binary or continuous). The unconditional weights are fixed:
\begin{equation}
W_i^{\mathrm{uncond}}(t)=W_{\text{init},i}\quad \forall i,t.
\label{eq:uncond_weights}
\end{equation}

\paragraph{Guided prediction.} Let \(\epsilon_{\mathrm{cond}}=\epsilon_\theta(x_t\mid \{W_i^{\mathrm{cond}}(t)\},\mathrm{cond})\) be the noise estimate using conditional weights and \(\epsilon_{\mathrm{uncond}}=\epsilon_\theta(x_t\mid \{W_i^{\mathrm{uncond}}(t)\},\varnothing)\) be the estimate using unconditional weights. The final ACFG prediction is:
\begin{equation}
\epsilon_{\theta}^{\mathrm{acfg}}(t)=(1+\omega)\,\epsilon_{\mathrm{cond}}-\omega\,\epsilon_{\mathrm{uncond}}.
\label{eq:acfg_prediction}
\end{equation}
where \(\omega\) is the guidance scale.

\begin{figure*}[htbp!]
    \centering
    \includegraphics[width=\textwidth]{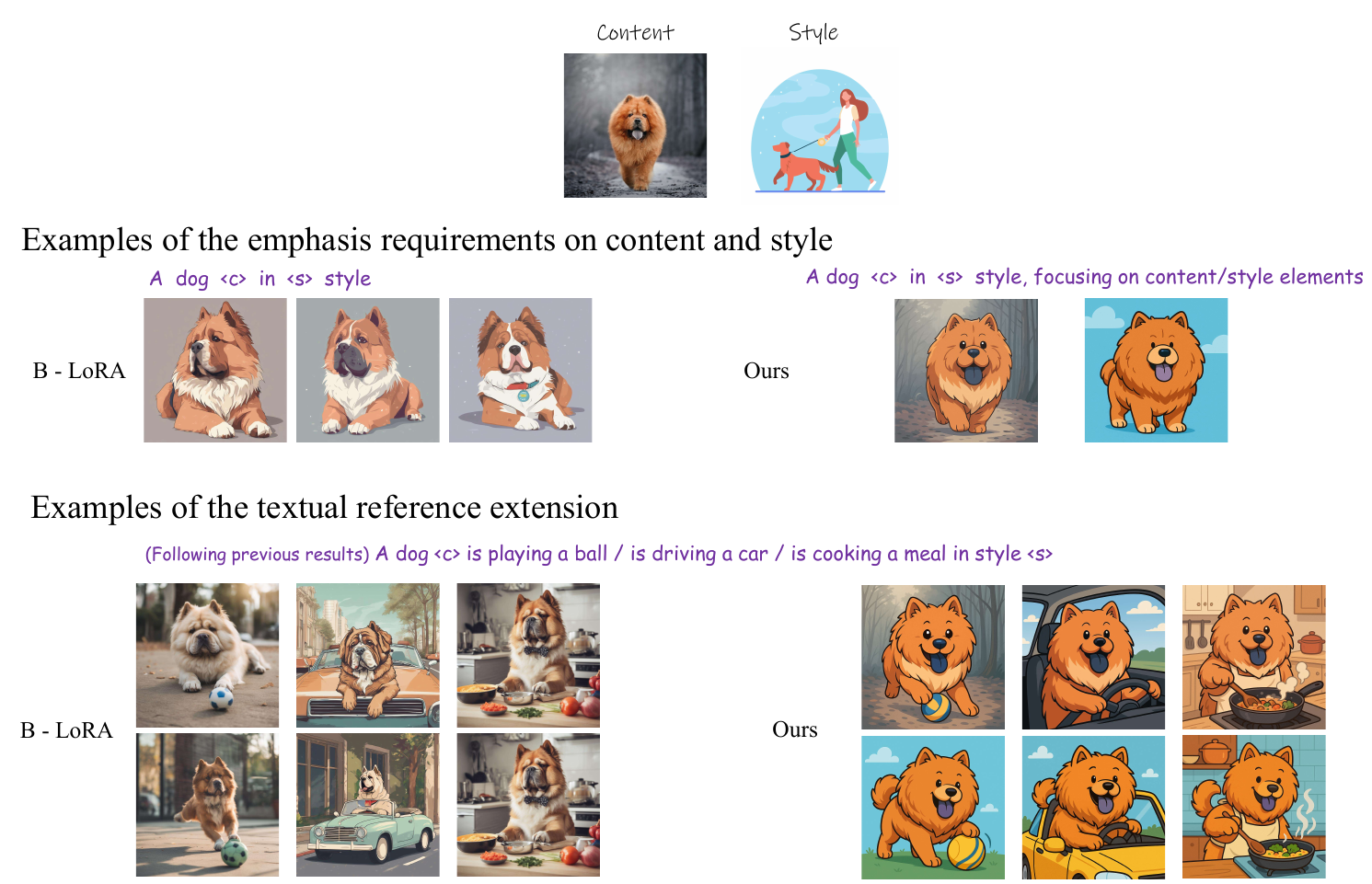}
    \caption{Additional visual results, including ACFG comparisons and semantic extensions.}
    \label{fig:visual_app2}
\end{figure*}

\paragraph{Zero-shot temporal scaling.} We propose a smooth schedule
\[
\alpha(t)=\alpha_{\min}+(\alpha_{\max}-\alpha_{\min})\,g\!\left(\frac{T-t}{T}\right),
\]
with a monotone function \(g:[0,1]\to[0,1]\) (e.g., a cosine schedule). This \(\alpha(t)\) modulates the aggregation weights \(\gamma_c(t)\) and \(\gamma_s(t)\) to emphasize content early in the sampling (high \(t\)) and refine style later (low \(t\)), while always keeping the unconditional branch fixed as in Eq.~\ref{eq:uncond_weights}.

\begin{table}[htbp!]
\centering
\caption{Performance by content–style pairing with ACFG. Metric shown is a composite quality score (higher is better).}
\begingroup
\setlength{\tabcolsep}{8pt}
\renewcommand{\arraystretch}{1.08}
\resizebox{\linewidth}{!}{
\begin{tabular}{@{}lccc@{}}
\toprule
\textbf{Content–style pairing} & \textbf{Fixed-weights baseline} & \textbf{ACFG (ours)} & \textbf{Relative improvement} \\
\midrule
Similar semantics & 0.81 & 0.87 & +7.4\% \\
Distant semantics & 0.68 & 0.79 & +16.2\% \\
Complex composition & 0.65 & 0.80 & +23.1\% \\
\bottomrule
\end{tabular}}
\endgroup
\label{tab:pairing_performance}
\end{table}

\begin{table}[htbp!]
\centering
\caption{Ablation of ACFG components. Higher indicates better quality.}
\begingroup
\setlength{\tabcolsep}{6pt}
\renewcommand{\arraystretch}{1.12}
\resizebox{\linewidth}{!}{
\begin{tabular}{@{}p{0.62\linewidth}cc@{}}
\toprule
\textbf{Component configuration} & \textbf{Quality score} & \textbf{Relative change} \\
\midrule
Full ACFG as proposed & 0.86 & Reference \\
Constant temporal scaling (no time dependence) & 0.78 & \(-9.3\%\) \\
No low-rank constraint on temporal adjustments & 0.82 & \(-4.7\%\) \\
Standard CFG (uncond. path uses cond. weights) & 0.75 & \(-12.8\%\) \\
\bottomrule
\end{tabular}}
\endgroup
\label{tab:ablation_study}
\end{table}

\section{Additional Visualizations}
\label{app:viz}
Figures~\ref{fig:visual_app1} and \ref{fig:visual_app2} provide additional qualitative results, complementing the visualizations in the main paper. These examples cover diverse objects, complex scenes, and challenging artistic styles, further demonstrating the robustness of our method. We also include failure cases to illustrate current limitations, such as difficulty with highly abstract styles or extreme compositional changes.

\end{document}